\documentclass[twoside,11pt]{article}

\usepackage{jmlr2e}

\usepackage{times}
\usepackage{latexsym}

\usepackage{graphicx}
\graphicspath{ {./images/} }
\usepackage{microtype}
\usepackage{multirow}
\usepackage{xcolor,colortbl}
\usepackage[colorlinks=true]{hyperref}
\definecolor{darkblue}{rgb}{0, 0, 0.5}
\hypersetup{linkcolor=darkblue,citecolor=darkblue}

\renewcommand\cite{\citep}
\firstpageno{1}

\begin{document}

\title{Unsupervised Automatic Speech Recognition : A Review\thanks{Speech Communication DOI: 10.1016/j.specom.2022.02.005}}

\author{\name Hanan Aldarmaki\thanks{Corresponding author.} \email h-aldarmaki@uaeu.ac.ae \\
       \addr Computer Science \& Software Engineering Department\\
       UAE University\\
       Al Ain, UAE
       \AND
       \name Asad Ullah \\ 
       \addr Department of Computer Engineering\\
       National University of Science \& Technology \\
       Islamabad, Pakistan
       \AND 
       \name Sreepratha Ram \\ 
       \addr Computer Science \& Software Engineering Department\\
       UAE University
       \AND 
       \name Nazar Zaki \\ 
       \addr Computer Science \& Software Engineering Department\\
       UAE University
       }

\maketitle

\begin{abstract}
Automatic Speech Recognition (ASR) systems can be trained to achieve remarkable performance given large amounts of manually transcribed speech, but large labeled data sets can be difficult or expensive to acquire for all languages of interest. In this paper, we review the research literature to identify models and ideas that could lead to fully unsupervised ASR, including unsupervised sub-word and word modeling, unsupervised segmentation of the speech signal, and unsupervised mapping from speech segments to text. The objective of the study is to identify the limitations of what can be learned from speech data alone and to understand the minimum requirements for speech recognition. Identifying these limitations would help optimize the resources and efforts in ASR development for low-resource languages. 
\end{abstract}

\section{Introduction}

What can be learned from a raw speech signal? This question has practical implications for low-resource Automatic Speech Recognition (ASR) and is also relevant for the study of human language acquisition. Modern ASR systems rely on large amounts of annotated speech to learn accurate speech representation and recognition, and they can achieve remarkable accuracy for resource-rich languages like English. At the time of writing, the state-of-the-art ASR model for English, which achieved 1.9\% word error rate on clean test data \cite{gulati2020conformer}, was trained using more than 900 hours of labeled speech. 
For a language like Arabic, which includes various dialects with some annotated resources, the word error rate using supervised methods is much higher: 13\% for standard Arabic, and around 40\% for dialects \cite{ali2017speech}. Many languages and dialects do not have any annotated resources or even a standard written form. Acquiring and labeling large datasets can certainly lead to better performance, but other factors could potentially be exploited to improve performance much more efficiently using the existing resources. 

We know that humans manage to acquire language without reliance on such massive resources or direct supervision---although other environmental and interactive cues certainly help since language is rarely used in isolation. Still, identifying what can be learned from the speech signal alone can illuminate some aspects of language acquisition on the one hand\footnote{For a review of computational models of language acquisition, see \citet{rasanen2012computational}.} and aid the construction of ASR systems for low-resource languages on the other. Our objective in this paper is to present relevant literature that can pave the way to unsupervised ASR : how to achieve reasonable ASR performance without acquiring labeled datasets. By assimilating various research efforts in this vein we hope that a clearer picture would emerge about the challenges presented by this task and promising directions for future work.

Supervised ASR models implicitly address various sub-problems without needing to explicitly model each one of them, as demonstrated in recent end-to-end neural models \cite{synnaeve2019end}. These sub-problems include segmentation, sub-word and word modeling, handling speaker and environmental variations, and classification into text labels.  In the absence of transcribed speech for supervision, each of these sub-problems presents a challenge that has to be addressed, often explicitly and sometimes independently of the other sub-tasks. The works we review in this paper are categorized according to the sub-task they attempt to address. Based on the wide range of works we studied, we outline a feasible framework for completely unsupervised ASR in figure \ref{fig:unsup_asr}.

\begin{figure}[h]
    \centering
    \includegraphics[width=16cm]{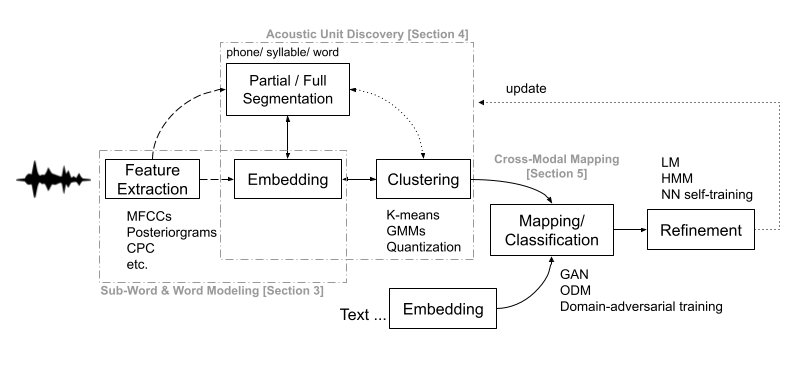}
    \caption{A high-level sketch of unsupervised ASR pipeline and possible sub-tasks.}
\end{figure} \label{fig:unsup_asr}

We summarize the process of unsupervised speech recognition as follows: given a raw signal corresponding to an utterance, we need to identify the meaningful units in the sound stream. This is the process of segmentation, which can be analyzed at multiple levels---phones, syllables, words, and collocations. Given the variable nature of speech, which arises from different speaker characteristics, environmental conditions, and other factors, we need to find suitable abstract representations of the raw speech segments to aid generalization. Learning features that are linguistically relevant and discriminative can be carried out at the frame and sub-word level (sub-word modeling) or word level (spoken word embeddings). These representations ideally summarize the phonetic and/or semantic content of each segment. In Figure \ref{fig:unsup_asr}, unsupervised sub-word and word modeling is shown as a combination of the feature extraction and embedding blocks, where features could be learned at the level of individual frames and/or longer segments. After segmentation and embedding, the segment embeddings should be clustered and classified into similar units---to identify unique word types, for instance. Identifying recurring patterns in the sound stream and clustering them into coherent units is what we call \textit{acoustic unit discovery}, which could be achieved using either full or partial segmentation and clustering. 
Unsupervised ASR could be tackled with various approaches, and we attempt to summarize the possible approaches in the figure by showing the alternative routes that could be taken, where some steps could be skipped, combined, or approached in an alternating manner; in particular, we find that segmentation, embedding, and clustering can be effectively modeled together instead of as separate processes. 

Unsupervised learning can only discover recurring patterns in the input signal and the relationships between those patterns. In speech, for example, the discovered patterns would not necessarily align with orthographically valid units like words in a dictionary. To obtain speech segments and clusters that are consistent with text, other modalities must be used for grounding. Supervised learning provides direct grounding by specifying the classification categories for each input unit. For example, in supervised ASR, each spoken utterance is paired with a text transcription. Such pairings, if available in abundance, enable end-to-end models to learn suitable embeddings and alignments between the speech and text in one go. Indirect grounding, or distant supervision, is the process of using a related but unaligned context of other modality (e.g. text or images) to ground the discovered patterns by finding correlations between the two cross-modal contexts. Using text for grounding, this  step ideally results in aligned speech and text segments, which can be used directly as a rudimentary ASR system by mapping each speech segment to its nearest neighbor in the text domain. Additional steps could be followed to refine the model and improve performance; for example, by incorporating a language model and Viterbi decoding, or using the initial labels as a noisy dataset for subsequent training in a pseudo-supervised manner. Using images with corresponding audio captions has also been explored for ASR-free image search, but also as an indirect grounding for spoken term discovery \cite{harwath2016unsupervised}. For the purpose of ASR, such models could potentially be used to eventually align the spoken words with text, but parallel image and captions should be available for both text and speech to make that possible.

\subsection{The Zero Resource Speech Challenge}

The Zero Resource Speech Challenge (ZRSC) was initiated in 2015 \cite{versteegh2015zero} with the goal of  accelerating research in the field of unsupervised speech processing. The end goal of this competition is to build a system that can achieve end-to-end learning of an unknown language, taking as input nothing but raw speech. The tasks handled by the challenge do not use any text for input or output, with the aim of building speech-only models, such as speech synthesis without text \cite{dunbar2019zero}, and language modeling without text \cite{dunbar2021interspeech}. The overarching objective of the challenge is only partially aligned with our formulation of unsupervised ASR, where we do in fact require text transcriptions as an output, so we do not restrict the use of un-aligned text for language modeling or other tasks. However, the challenge does address some sub-tasks that are useful for unsupervised ASR, including subword modeling, spoken term discovery and word segmentation.  We review relevant models that were submitted to the challenge throughout the paper. In particular, we focus on speaker-invariant sub-word modeling, were the goal is to attain robust representations of speech sounds that ideally encode the relevant linguistic features and discard irrelevant acoustic features, such as speaker characteristics. In addition, we review works on acoustic unit discovery, which aims to identify recurring patterns in spoken utterances. This could entail partial or full segmentation of utterances into smaller units. More details about the challenge and the submissions for each task can be found in the the challenge's main review papers \cite{versteegh2015zero,dunbar2017zero,dunbar2019zero,dunbar2020zero,dunbar2021interspeech}.

\subsection{Scope}

For this review, we assimilated relevant research literature in the following subareas: unsupervised sub-word and word modeling (section \ref{sec:rep}), unsupervised segmentation and spoken term discovery (section \ref{sec:seg}), and cross-modal mapping (section \ref{sec:dist}). 
For the purpose of presenting an insightful and coherent discussion, we did not limit the time frame of the discussed literature; the main criterion of inclusion is relevance and impact of the research outcomes on subsequent research efforts. For brevity, we do not include details of models that have been discussed and compared in existing reviews and provide citations for further reading instead. 

In unsupervised sub-word and word modeling, we include works on sub-word modeling as defined in the Zero Resource Speech Challenge (section \ref{sec:sub}), and other major and recent works on unsupervised acoustic word embeddings (section \ref{sec:emb}). For spoken term discovery, we discuss models that concretely aim to discover recurring patterns that ideally correspond to words. This sub-task overlaps with word segmentation as some segmentation models take the extra step of clustering the segments to identify recurring units. However, we keep the discussion of the two subtasks in separate sections since full segmentation models have a somewhat different objective, which is to identify word boundaries. After segmentation, we discuss approaches for mapping the segments to textual units using distant supervision, where an independent text corpus is used for cross-modal mapping. Approaches in this category include several recent efforts that utilize adversarial networks for unsupervised mapping with moderate to high success. 

To get a complete picture and better understanding of these unsupervised models in the context of modern ASR, we start by describing the components of standard and state-of-the-art ASR systems in section \ref{sec:back}. 


\subsection{Terminology}

In this paper, we address the problem of unsupervised Automatic Speech Recognition (ASR), which in this context refers to the problem of generating text transcriptions from raw speech input. By ``supervision", we mean any form of manual labeling generated by humans, such as pre-transcribed spoken utterances, phone and word boundaries, and pronunciation dictionaries. Therefore, any model that does not utilize such resources is unsupervised by our definition. We include self-supervised models that use an auxiliary supervised objective from the unlabeled input itself (such as auto-encoders) in our definition of unsupervised ASR. We also include models that utilize non-parallel resources of other modality, particularly text. 

In speech science, a ``phoneme" is the smallest unit that distinguishes a word from another in a given language. However, phonemes do not necessarily correspond to coherent acoustic units, and they are language-dependent \cite{moore2019use}. Acoustic units in that range are referred to as ``phones". Transcriptions of speech could be either ``phonetic", representing the sounds actually present in a given utterance, or ``phonemic", representing an abstract and consistent form of each word in the language. Acoustic models typically model phones and eventually classify them into phonemes. Instances of these terms in the rest of the paper should be interpreted according to these definitions. 

\section{Background}
\label{sec:back}
Automatic Speech Recognition (ASR) is the process of automatically identifying patterns in a speech waveform. Patterns that could be detected from speech include the speaker's identity, language, emotion and the textual transcription of the spoken utterance. The latter is what is typically sought in ASR and is the focus of this paper.

The smallest recognizable unit of speech is the phoneme, which are the sounds that distinguish words in a given language. An acoustic realization of a phoneme in actual utterances is called a phone, with a duration of 80 ms on average with high variance from 10 to 200 ms. Phones are produced by changes in the shape of the speaker's vocal tract (VT), and spectral patterns of the speech signal indirectly encode these VT shapes \cite{Shaughnessy}. Sequences of phones compose words and utterances that carry meaning. 

\subsection{Traditional ASR} 
Typical ASR models are composed of three main components: an acoustic model, a pronunciation dictionary and a language model.

The Acoustic Model (AM) calculates the probability of acoustic units (e.g phones, sub-word units etc.), which can be modeled using Gaussian Mixture Models (GMMs) \cite{389219} and Hidden Markov Models (HMMs) \cite{10.2307/1268779}. Typically, GMMs are used to compute the probability distribution of phones in a single state while HMMs are used to find the transition probability from one state to another. 
Each state corresponds to an acoustic event, such as a phone. The GMM-HMM model is trained by the expectation maximization (EM) technique, and Viterbi decoding is used to find the optimal state sequence in HMMs. The pronunciation of phones in natural utterances often varies depending on the acoustic context; therefore, context-dependent triphone HMMs are used to model speech sounds, where each phone is modeled with a left and right context.
Recent ASR models have replaced GMMs with deep neural networks (DNNs) \cite{hinton}. These models are dubbed hybrid DNN-HNN models, and they are still widely used as competitive ASR models. 

The Language Model (LM) component computes the probability of a sequence of words. LMs are used to improve the accuracy of acoustic models by incorporating linguistic knowledge from large text corpora. Syntactic and semantic rules are learned implicitly in LMs, which are then used to re-score the acoustic model hypotheses. To align the phonetic transcriptions that result from the AM with the raw text used in language models, a pronunciation dictionary is used to map a sequence of phonemes into words.

\begin{figure}[htp]
    \centering
    \includegraphics[width=13cm]{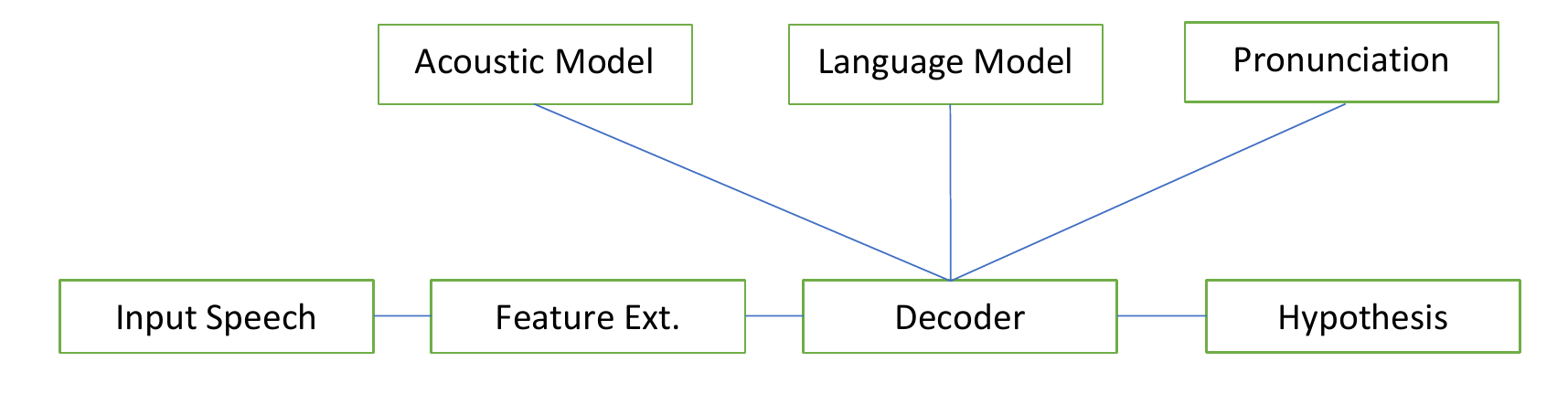}
    \caption{Traditional ASR Pipeline}\label{Fig:1}
\end{figure}

These components are trained independently and combined to build a search graph using Finite state transducers (FSTs). The decoder then generates lattices that are scored and ranked to generate the target word sequences. ASR models are typically evaluated using word error rate (WER), which is the number of substitutions, insertions, and deletions, divided by the total number of words in the target transcription. The phone error rate (PER) is another metric used to measure the performance of the acoustic model. A block diagram of Traditional ASR is shown in Figure \ref{Fig:1}.

\subsection{Modern ASR}  \label{sec:asr}

Modern ASR systems are fully end-to-end; for example, \citet{amodei2015deep} describes an encoder-decoder architecture, where the input audio is processed using a cascade of convolutional layers to produce a compact vector. The decoder then takes the encoded vector as input and generates a sequence of characters. A number of different objective functions such as CTC \cite{graves}, ASG \cite{Collobert2016Wav2LetterAE}, LF-MMI \cite{Hadian2018EndtoendSR}, sequence-to-sequence \cite{Chiu2018StateoftheArtSR}, Transduction \cite{Prabhavalkar2017ACO} and Differentiable decoding \cite{pmlr-v97-collobert19a} can be used to optimize end-to-end ASR. Moreover, different architectures of neural networks such as ResNet \cite{He2016DeepRL}, TDS \cite{Hannun2019SequencetoSequenceSR} and Transformer \cite{Vaswani2017AttentionIA} have been explored. The output labels of the end-to-end system can be characters or subword units such as byte-pair encoding (BPE). An external LM can be incorporated to improve the overall system performance. 

\begin{figure}[htp]
    \centering
    \includegraphics[width=13cm]{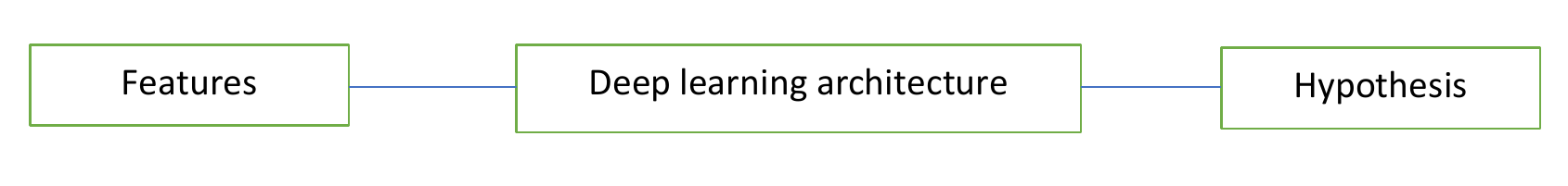}
    \caption{Modern ASR Pipeline}\label{Fig:2}
\end{figure}

The end-to-end ASR pipeline is shown in Figure \ref{Fig:2}.

\subsection{Challenges in ASR} 

One of the challenges in ASR, even if supervised, is the variability that is characteristic of natural spoken utterances due to speaker and environmental conditions. 
Utterances by different speakers have acoustic differences that can be difficult to disentangle from the phonetic content. Even for an individual speaker, variability arises due to speaking rate, intensity, affect, etc. 
Furthermore, ASR models are often trained on clean speech data but they are often evaluated on real-time noisy speech data. Sources of noise include background noises and signal distortions through the input device.

Speaker-independent models could be trained using data that includes multiple speakers, but this often degrades the performance of ASR and requires larger amounts of data for training to achieve decent performance. The same applies to background noises and different environmental conditions. Instead, state-of-the-art ASR systems are speaker-adaptive: they capture the variability of speakers using I-Vectors \cite{6707705} and X-Vectors \cite{8461375} which are low-dimensional vectors that encode speaker-specific features. In addition, various augmentation techniques can be utilized to supplement the training data with more examples that reflect the expected variability in test conditions.

For example, volume and speed perturbation are used to capture the variability between utterances. Similarly, noise-augmentation is used to supplement the training data with different environmental conditions \cite{Ko2015AudioAF}. All these strategies are combined to train robust multi-condition ASR systems that can handle multiple sources of variability.

\subsection{Feature Extraction}\label{sec:mfcc}

The first step in any ASR pipeline is feature extraction, i.e. extracting meaningful information from speech and discarding redundant information. The power spectrum of the speech signal somehow encodes the shape of the vocal tract, which determines the sound (phone) generated, in addition to other speaker-specific characteristics. The most commonly used representation of the speech power spectrum is the Mel-frequency cepstral coefficients (MFCCs), which are widely used as the very first step in speech processing to convert the audio signal into discrete frames.  

\begin{figure}[h]
    \centering
    \includegraphics[width=7.5cm]{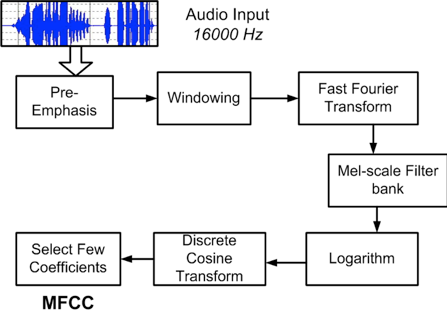}
    \caption{MFCC feature extraction}\cite{asadullah}
\end{figure}

A typical application of MFCC feature extraction is carried out as follows: the signal is first pre-emphasized to amplify the high frequency components, then the signal is segmented into uniform overlapping frames, typically 20 ms in duration. 
The frames are pre-processed by a hamming window function before applying Fast Fourier Transform (FFT) to compute the power spectrum. The output is passed through $\sim$25 Filter banks to get the spectrogram of the speech signal, which can be used directly as the input vector in ASR, as done in some end-to-end neural models. However, these features are high-dimensional and highly correlated. Alternatively, the Discrete Cosine Transform (DCT) of the log filter bank energies is calculated, and the first 13 DCT coefficients are selected as compressed and decorrelated features. Additional features include 13 delta-coefficients and 13 acceleration coefficients derived from the selected DCT coefficients, which can be combined to make 39 dimensional vectors. The whole process is shown in Figure 3. While these MFCC features efficiently encode useful phonetic features, they also encode other acoustic features like speaker and environmental conditions. 

An alternative feature representation is perceptual linear predictive (PLP) introduced in \citet{hermansky1990perceptual}, which is found to be more robust to speaker variations. Additional robustness could also be gained using Gaussian posteriorgrams, which are obtained from the frame-wise posterior probabilities of each phonetic class using Gaussian Mixture Models (GMMs) trained on MFCC or PLP features \cite{hazen2009query}.

\section{Acoustic Sub-Word \& Word Modeling}\label{sec:rep}

Most ASR models start by extracting salient features from the raw waveform. 
The purpose of an initial feature extraction is to downplay any linguistically-irrelevant patterns in the waveform that are likely to distract from the learning task at hand, such as speaker characteristics, channel distortions, environment noise, etc. As described in section \ref{sec:mfcc} above, MFCCs are widely used features in most ASR applications, and many of the approaches discussed here actually start with MFCC features before performing additional feature modeling and embedding steps. The downside of MFCCs is that they still contain speaker-related features; using additional transformations and feature embedding can lead to more robust representations, particularly in unsupervised settings. 

We divide this section into two parts: (\ref{sec:sub}) unsupervised sub-word modeling, which is mainly derived from the Zero Resource Speech Challenge (ZRSC) and focuses on speaker-invariant frame-level feature representations, and (\ref{sec:emb}) acoustic word embeddings, where fixed-length embeddings for longer segments are obtained. Note that acoustic word embedding models could work either with MFCC features or any feature transformation described in the first subsection. Also, it's worth noting that the acoustic word embedding models themselves could be seen as a form of subword modeling, since they can potentially by applied on any level of processing, including individual frames.

\subsection{Speaker-Invariant Sub-Word Modeling}\label{sec:sub}

Subword modeling involves learning speech features that are linguistically relevant (i.e. features that discriminate between different phone categories) while discarding irrelevant acoustic features that do not contain linguistic information. Ideally, the learned features should be robust to speaker variations, and they may even generalize across different languages \cite{dunbar2017zero}. The standard metric used for evaluation is the ABX discriminability between phonemic minimal pairs \cite{schatz2013evaluating}, which measures the ability of the model to recognize instances of the same phoneme in different contexts and different speakers. The ABX metric uses three tokens, A and B, which differ by one phoneme (e.g. b-a-g vs. b-e-g), and a third token X, that could belong either to the same category as A or B. Models are evaluated by classifying X to either A or B based on some similarity metric (such as dynamic time warping and frame-level cosine similarity), and averaging the results over all ABX pairs in the test set. To evaluate speaker-invariance, test sets are constructed such that A and B belong to one speaker, while X belongs to a different speaker.

\par

Clustering can be used as a simple form of feature representation \cite{coates2012learning}, and frame-level clustering has been proposed for speaker-invariant subword modeling using Gaussian posteriorgrams \cite{zhang2010towards}. The most commonly used clustering approach for this task is a Dirichlet Process Gaussian Mixture Model (DPGMM), as done in \citet{chen2015parallel} and \citet{heck2016unsupervised}. K-means has also been used successfully for subword modeling in \citet{pellegrini2017technical}.

Due to the high sensitivity of the DPGMMs to acoustic variations, a DPGMM produces too many classes leading to very high dimensional posteriorgrams. To address this issue, \citet{heck2016unsupervised} used Linear discriminant analysis (LDA) to transform speech vectors before clustering. \citet{heck2017feature} extended this idea to include additional feature transformations: Maximum likelihood linear transforms (MLLT), Feature-space maximum likelihood linear regression(fMLLR) and basis fMLLR transformations. 
Frame-level DPGMMs are learned separately for each set of transformations, and the combination of their posteriorgrams is used as the final feature representation for each frame, which led to better performance compared to using raw features or individual transformations. The performance of the models proposed by \citet{heck2017feature}  surpassed all other models submitted in ZRSC'17 in ABX evaluations. 
It's worth noting that they have used PLP features \cite{hermansky1990perceptual} instead of MFCCs as input, which can be partially responsible for the superior performance in cross-speaker ABX evaluation.

\citet{pellegrini2017technical} uses K-means clustering on MFCC features whitened using Zero-Component Analysis (ZCA). 
The feature representations are calculated as the distances between data points and the cluster centroids. While this model outperforms the baseline of using MFCC features alone, it considerably underperforms compared with DPGMM-based models.

An alternative use of clustering for sub-word modeling is to use the automatically assigned cluster labels to form an auxiliary supervised training target for training neural network models, as done in \citet{pellegrini2017technical} and other submissions in ZRSC'17, such as \citet{yuan2017pairwise}, \citet{yuan2016learning}, and \citet{chen2017multilingual}. However, using neural networks in this manner did not lead to better performance compared to simply using the features derived directly from the clustering methods. Earlier models described in \citet{badino2014auto,badino2015discovering} use variants of deep Auto-Encoders (AEs) for sub-word modeling, and show that AEs lead to more discriminative features compared to GMMs.  They also experiment with intermediate binarized features that are used for both encoding and clustering while minimizing the reconstruction loss. However, standard AEs generally perform better in ABX evaluations \cite{badino2015discovering}.  

More recent models that were evaluated within the context of speech synthesis without text \cite{dunbar2019zero} resulted in much better performance in ABX evaluations. The best-performing models in this task use Contrastive Predictive Coding (CPC),  which helps representations to capture phonetic contrasts \cite{riviere2020unsupervised, kahn2020libri}. A good example of a model that follows this framework is described in \citet{van2020vector}.

In addition to using CPC for feature representation, \citet{van2020vector} explores the use of vector quantized neural networks in conjunction with autoencoders. Vector quantization works by mapping the continuous feature vectors to their nearest neighbour in a codebook containing a finite number of distinct codes (or features), thereby effectively discretizing the features.
In particular, they use vector quantized variational autoencoder(VQ-VAE) for acoustic modelling. A vector quantization layer is inserted between the encoder and decoder to discretize the representations learned by the encoder. In order to verify that VQ is robust to speaker variations, the representations are evaluated before and after quantization. This is verified by a clear reduction in speaker classification accuracy post vector quantization. 

\citet{gundogdu2020vector} also applys vector quantization to recurrent sparse autoencoders, which are fine-tuned using the respective speech segments obtained using unsupervised term discovery (as described in \citet{yusuf2019temporally}). This system performs almost as well as the topline in ZRSC'20 in ABX tests for training languages, and surpasses the topline for surprise languages. Similarly, \citet{tobing2020cyclic} applies vector quantization to cyclic VAEs (CycleVAE) to improve performance by detangling speaker characteristics from the latent space. This is achieved by marginalizing possible speaker conversion pairs. While CycleVAE performs worse than the baseline in ZRSC'20,  vector quantized CycleVAE performs almost as well as the topline, indicating the positive influence of vector quantization on learning better subword representations.

One of the drawbacks of using a VQ-VAE model is that this model encoded speech in much smaller segments (higher bitrate) than human transcriptions of phonemes. To address this issue, \citet{kamper2020towards} proposes to match blocks of contiguous features vectors to a single codebook entry (rather than matching each feature to a code). Two methods are experimented with for segmentation of these feature blocks: a greedy approach and a dynamic programming approach. The greedy approach merges a predefined number of feature vectors, while the dynamic programming approach optimizes the sum of squared distances between feature vectors and associated codes within each segment. This essentially amounts to phonetic segmentation, subword modeling, and acoustic unit discovery. Additional models that utilize quantization in the context of speech synthesis without text are described in \citet{dunbar2020zero}, where models are evaluated for subword modeling and term discovery, in addition to speech synthesis quality.  
\par

\subsection{Acoustic Word Embeddings} \label{sec:emb}

Variability in speech segment length makes it difficult to directly compare frame-wise representations of acoustic units like phones or words. Dynamic Time Warping (DTW) is an early technique used to compare variable-length audio segments using dynamic programming to find an optimal frame-wise alignment between them \cite{rabiner1978considerations, 4156191, levin2013fixed}. DTW is based on frame-wise comparison, typically using the Euclidean distance between each pair of frames, so it is used mostly to compare segments representing the same acoustic unit; for example, the same word spoken at different rates. Furthermore, since MFCCs include a speaker and environmental characteristics in addition to phonetic features, MFCC-based DTW cannot effectively be used to compare segments with non-matching conditions.  Using posterior features can lead to more robust performance \cite{aradilla2006using,hazen2009query}. 

Computationally efficient models attempt to embed the segments into fixed-dimensional vectors that can be directly compared using Euclidean distance or cosine similarity, which enables scalable spoken term discovery \cite{Park2008UnsupervisedPD,inproceedings,6163965} and query-by-example search \cite{article,5372931,inproceedings2}. 
Early embedding approaches include simple down-sampling \cite{10.3115/100964.100983,article:glass,Ostendorf95fromhmms,article:abdel}, acoustic model-derived features \cite{inproceedings:zweg,article:layton} and convolutional deep neural networks \cite{Maas2012WordlevelAM}.

Down-sampling is a rather simple technique to embed segments directly by extracting a specified number of frames from each segment. For instance, uniform down-sampling is performed by sampling frames at T/k intervals, where T is the total number of frames in the segment and k is the number of samples we wish to extract. The resultant embedding size is then k times the dimension of the MFCC feature vectors. Non-uniform down-sampling can also be achieved using a k-state HMM, where each state is modeled as a single spherical Gaussian \cite{levin2013fixed}. Variants of unifrom and non-uniform downsampling strategies are explored in \citet{holzenberger2018learning}.


Alternatively, neural networks can be used to compress variable-sized input segments into fixed-dimensional embeddings, as is typically done for text-based embeddings such as word2vec \cite{inproceedings:mikolov}, which is used to obtain continuous vector representations of words such that semantically or syntactically related words are similar to each other in the vector space. These embeddings can be trained using the continuous bag of words (CBOW) or skipgram objectives. In CBOW, the model attempts to predict a target word given a window of surrounding context words, while in skipgram, the model directly minimizes the distance between the embedding of a word and selected context words, with negative samples to regularize the training. 

For spoken words, the embedding network must somehow handle the variable-length input of acoustic frames. The most common architecture for this type of acoustic embedding is a neural encoder-decoder self-supervised model.

For example, audio word2vec \cite{chung2016audio} and speech2vec \cite{chung2018speech2vec} are both directly inspired by the word2vec text embedding framework to obtain spoken word embeddings, but they rely on recurrent encoders and decoders to handle the variable-length nature of spoken words. Audio word2vec's training objective is not parallel to word2vec's CBOW or skipgram objectives. Each segment is encoded and decoded independently of other segments; as a result, the obtained embeddings encode phonetic rather than semantic features, which is useful in cases where acoustic similarity is more desirable than semantic similarity. On the other hand, speech2vec \cite{chung2018speech2vec} employs a training methodology borrowed directly from the word2vec framework using the skipgram objective. Since each occurrence of a spoken term is rather unique, static speech embeddings can be obtained by averaging the embeddings of all occurrences of a spoken word. Note that unlike text-based skipgram where negative samples are necessary to avoid a degenerative solution, speech2vec does not require such regularization since the decoder has to reconstruct the original segment frame-by-frame, thereby ensuring that different words have different embeddings. Since they emphasize similarity to context words, speech2vec embeddings tend to encode semantic features rather than phonetic features as most other speech embedding techniques. 

Similar RNN-based auto-encoders for acoustic word embeddings are described in \citet{audhkhasi2017end} and \citet{holzenberger2018learning}, the latter shows that RNN-based models outperform down-sampling in ABX evaluations.

A more robust form of self-supervised learning of acoustic embeddings is the Correspondence Auto-Encoder (CAE) described in \citet{kamper2019truly}, where the decoder is trained to reconstruct another occurrence of the spoken word instead of directly reconstructing the input as in audio word2vec, or surrounding context as in speech2vec. Pairs of spoken words are first collected using an unsupervised term discovery system, and the CAE model is then used to reconstruct one of the words given the other as input. 
The model was evaluated on word discrimination tasks designed in \citet{carlin2011rapid}, and it's compared against downsampling, auto-encoders, and variational auto-encoders. CAE embeddings outperformed all others in this evaluation task.  
However, a correspondence variational auto-encoder is later introduced  \cite{peng2020correspondence}, where it outperforms the  original EncDec-CAE model in word discrimination tasks. 

Acoustic word embeddings could be improved further using pre-trained frame-wise features instead of directly using MFCC vectors as input to the embedding network, as shown in \citet{van2021comparison}.  They show that using frame-wise features that are trained using the following approaches lead to better acoustic word embeddings than using MFCC features directly: contrastive predictive coding (CPC) \cite{van2018representation}, Autoregressive predictive coding (APC) \cite{chung2019unsupervised}, and a frame-level correspondence autoencoder (CAE). Similar to \cite{kamper2019truly}, the latter is trained by first extracting pairs using unsupervised spoken term discovery, then DTW is used to find a frame-level alignment between these words. The model is then trained at the frame level similar to a regular auto-encoder objective. 
These three frame-level feature representations (CPC, APC, or CAE) were used as input to an acoustic word embedding model, namely the correspondence auto-encoder model described above, and they all led to significantly better word discrimination accuracy when combined with downsampling and correspondence auto-encoders compared to using MFCC features. The improvement is particularly evident for the low-resource language Xintonga. Results on speaker classification accuracy indicate that such representations downplay speaker-specific features in favor of more linguistically-relevant features.
Not that all models described above could be applied to true word pairs with oracle boundaries, or pairs extracted in an unsupervised fashion. 
The audio word2vec and speech2vec models described above have been evaluated using oracle word boundaries, but they could theoretically be applied after unsupervised term discovery (section \ref{sec:term}) or full coverage segmentation (section \ref{sec:seg}), where phone/word boundaries are detected automatically in an unsupervised manner.

\section{Unsupervised Segmentation \& Acoustic Unit Discovery} \label{sec:seg}

Segmentation is the process of breaking down a continuous stream into discrete units, such as phones, syllables, words, or other meaningful sub-word units. These segments could then be processed, clustered, and used to efficiently access the information content of the utterance. 

Earlier works on lexical perception attempted to segment phonemically transcribed inputs of child-directed speech \cite{macwhinney1990child}. This data set was further processed using a phonemic dictionary so that each occurrence of a word type has the same phonemic transcription \cite{brent1999efficient}. This simplified formulation is not representative of the actual challenges in raw speech segmentation where phones are not known in advance or even perfectly segmented. However, the approaches and results described in these earlier works demonstrate the effectiveness of different linguistic assumptions (section \ref{sec:seg1}). Moreover, these approaches can be used in a bottom-up fashion after phonetic segmentation and clustering to identify possible locations of word boundaries (as explored, for example, in \cite{jansen2013summary} ). 

Identifying word boundaries from the raw speech is considerably more challenging. Phonetic (section \ref{sec:phone}) or syllabic (section \ref{sec:syl}) segmentation could be used as the first steps to constrain the locations of word boundaries. Phones and syllables have relatively predictable temporal structures that can be identified in the signal. However, clustering these phonetic or syllabic segments into types consistent with their true identities is more challenging and is often addressed within a distant supervision framework (see section \ref{sec:dist}). 

For unsupervised word segmentation of raw speech (Section \ref{sec:word}), several self-supervised models have been proposed. Most of these models are based on minimizing reconstruction loss in an auto-encoder framework. Some of these models rely on the assumption of within-word predictability and ignore wider context, while other models incorporate hierarchical structures more explicitly. %

Before describing full segmentation models, it's worth noting that earlier models focused on identifying occurrences of individual words, a task often dubbed ``spoken term discovery''. Compared to full-coverage segmentation, these methods could potentially be more accurate in identifying frequent words and can be useful as a first step for full-coverage segmentation or in query-by-example search \cite{article,5372931,inproceedings2}. Full-coverage word segmentation (section \ref{sec:word}), on the other hand, results in complete segmentation of the input corpus without necessarily identifying recurring terms. Some approaches incorporate clustering and embedding with the full segmentation process to achieve rough word discovery in addition to segmentation. We will start by discussing models that directly address spoken term discovery in section \ref{sec:term}, followed by full segmentation models in the remaining sections. For the latter, we will start by reviewing models of word segmentation from phonemic transcriptions, followed by segmentation models from raw speech. Segmentation performance is reported using token F1-score (TF) and boundary F1-score (BF). The former counts only the segments where both boundaries are correctly detected and no spurious internal boundaries are added; the latter measures the detection of individual boundaries. In some cases, the R-value is reported as an alternative to the F1 measure  \cite{rasanen2009improved}, which is more robust in cases of over-segmentation (high recall and low precision). In raw speech segmentation, a tolerance of 20 ms is used in all measures; i.e. a boundary detected within 20 ms of a true boundary is considered correct.

\subsection{Spoken Term Discovery}\label{sec:term}

Several spoken term discovery models have been discussed and compared in three iterations of the Zero Resource Speech challenge \cite{versteegh2015zero, dunbar2017zero,dunbar2020zero}. In this section, we review a selection of models that showcase the major directions used for this task; additional models and details can be found in the ZRSC papers cited above.

Segmental dynamic time warping (S-DTW) \cite{park2008unsupervised} is used to find acoustically similar segments in audio utterances. The original DTW algorithm is best suited for aligning isolated word segments (see section \ref{sec:emb}. S-DTW is applied at the utterance level to find potential recurring patterns within these utterances. The DTW algorithm is modified by incorporating some constraints that limit the temporal skew of the alignment and offset starting points for the search, which results in multiple possible alignments for each pair of utterances. These constraints naturally divide the input into regions where the traditional DTW algorithm can be used to find the optimal alignment. The next step is to discard all alignments with high distortion and only keep the best matches. This is achieved by identifying segments with length at least L that minimize the average distortion \footnote{The distortion values are calculated using the Euclidean distances between the aligned acoustic vectors.}, which can be calculated in O(N log(L)) time \cite{lin2002efficient}, where N is the length of the fragment, and L is the minimum length of resulting subsequence. The minimum length L can be tuned to return linguistically meaningful units like words and phrases\footnote{In \cite{lin2002efficient}, L is set to 500 ms}. 
The discovered segments are then clustered using a graph clustering algorithm to identify unique word types.
 The nodes in the graph represent time locations in the audio stream that have frequent overlap with other points in the stream. The edges represent the similarity based on the average distortion score for the path common between the two nodes. An efficient clustering algorithm is then used to identify groups of similar nodes \cite{newman2004fast}. 
Clusters generated from 1-hr speech by the same speaker have high purity and good coverage of terms recurring in the sound stream. However, the approach does not provide full coverage as it relies on repeated patterns by the same speaker; the segments have to be acoustically similar. The approach relies on consistent recurring occurrence of speech patterns given similar speaker and environmental conditions. 
A probabilistic approach to DTW-based spoken term discovery is described in \citet{rasanen2020unsupervised}. 

In \citet{zhang2010towards}, the S-DTW algorithm is extended by using Gaussian posteriogram representation of the speech signal instead of MFCCs to generalize the approach for multiple-speakers. The approach is based on training an unsupervised GMM on speech segments from multiple speakers, and then using the trained GMM to generate the posterior vector for each input frame. These vectors are used for the next two steps of S-DTW and clustering. The distance metric used in this case is the negative joint log-likelihood, 
which is equivalent to the probability of the two vectors being drawn from the same underlying distribution. Experiments on the TIMIT dataset, which includes a total of 580 speakers, indicate that Gaussian Posteriograms can identify a much larger number of word clusters spanning multiple speakers with high cluster purity based on word identity. However, the same words were sometimes broken into different clusters. 

A model that integrates hierarchical levels of segmentation is the one described in \citet{lee2015unsupervised}. It combines the phonetic segmentation of \citet{lee2012nonparametric} with the adaptor grammar of  \citet{johnson2009improving} for word discovery. The adaptor grammar incorporates words, subwords, and phones, but it does not include collocations. They also model phone variability using a noisy-channel model to map the variable segments into unique types, which circumvents the need for explicit clustering. The noisy-channel, which attempts to standardize the phonetic segments, is implemented as a PCFG that includes three edit operations: substitute, split, and delete.  These three components (phonetic segmentation, variability modeling, and lexical segmentation) are modeled jointly as a generative process. Compared to other approaches, this joint model results in a higher word discovery rate when evaluated in a subset of six lectures from the MIT lecture corpus \cite{glass2005spoken}. The results are shown in Table 3, which is the average number of hits (discovered words) from a list of 20 frequent words. In addition, this model can be used for full-coverage phoneme and word segmentation. In this dataset, the model achieved 76 phoneme segmentation F score, and 18.6 word segmentation F score, averaged over the six lectures.

\begin{table*}[]
    \centering
    \begin{tabular}{|l|l|c|}
       \hline
       Model    &   Description &  Hit $\backslash$ 20  \\
       \hline
       \citet{park2008unsupervised} & Segmental DTW with MFCC features & 14.8\\       \cite{zhang2010towards} & Segmental DTW with Gaussian Posteriograms & 17.2\\
       \citet{lee2015unsupervised} A & Integrated acoustic model, noisy-channel \& adaptor grammar & 17.8\\
       \citet{lee2015unsupervised} B  & \citet{lee2015unsupervised} A, without acoustic model & 16.3\\
       \citet{lee2015unsupervised} C & \citet{lee2015unsupervised} A, without noisy channel & 11.5 \\
       \hline
    \end{tabular}
    \caption{Spoken term discovery results as hits over 20, or number of discovered words from a list of 20 words with top tfidf scores \cite{park2008unsupervised}. Results are reproduced from \cite{lee2015unsupervised}, but we report the average over the six lectures.}
    \label{tab:my_label}
\end{table*}


\subsection{Word segmentation from Phonetic Transcriptions}  \label{sec:seg1}

Statistical cues, such as the internal consistency of words, could be utilized for word segmentation \cite{saffran1996word}.  
Assuming that syllables and phonemes are more predictable within words than across word boundaries, transitional probabilities or mutual information were used in earlier models for word segmentation given phonemically transcribed speech \cite{cairns1997bootstrapping}. According to this view, given a unit (phone, syllable) naturally occurring in a spoken utterance, the likelihood of the next unit should be higher if both units form a word than if they cross word boundaries. The phoneme transitional probabilities, calculated from data, can thus be used to insert word boundaries at points of low probability. These models ignore word transition probabilities, assuming that words in an utterance are independent. 

The above assumptions have been implemented using standard n-gram modeling \cite{cairns1997bootstrapping} and self-supervised neural networks \cite{cairns1997bootstrapping, elman1990finding}. Using a self-supervised Simple Recurrent Network (SRN) with the objective of predicting the next phoneme, peaks in prediction errors are used as indicators of word boundaries. This model tends to over-segment the input, leading to units that are more like syllables than words \cite{cairns1997bootstrapping}. 

Other segmentation models based on the assumption of within-word predictability employ probabilistic word grammars, text compression schemes, minimum description-length\footnote{See \citet{brent1999efficient} for a complete review of these methods.}, and generative probabilistic models of unigram word distribution \cite{brent1999efficient}. Brent's Model-Based Dynamic Programming (MBDP) model is specified in a way that assigns a higher prior probability to segmentations with fewer and shorter lexical items by explicitly modeling relative frequencies.

The models described above ignore all syntactic relationships between words in an utterance. Yet syntax clearly plays a role in lexical perception \cite{rasanen2012computational}. While direct incorporation of syntactic structure in word segmentation has not been well studied, simpler distributional cues, such as word dependencies, can be used to indirectly incorporate syntax in the segmentation process. \citet{goldwater2006contextual} shows that incorporating context in the form of bigram dependencies leads to improved word segmentation performance. 
Unigram models tend to under-segment utterances by mistaking collocations\footnote{words that frequently occur together.} for words. Models that explicitly incorporate collocations can largely rectify this problem.%

A more detailed discussion can be found in \citet{goldwater2009bayesian}, which evaluates non-parametric Bayesian models that incorporate different independence assumptions. Assuming words are independent of each other, models tend to under-segment the utterance resulting in multi-word segments. \citet{goldwater2009bayesian} argues that this weakness is general for all models that assume unigram word distribution. Introducing dependencies between words increases the segmentation rate and accuracy. As in \citet{brent1999efficient}, \citet{goldwater2009bayesian} uses a probabilistic generative process to compute the prior probability of each possible segmentation. The probabilistic model is designed in a way that generates novel lexical items with high probability early in the process, and this probability decreases as more tokens are generated. This leads to fewer lexical items overall. Moreover, the probability of each novel word is the product of the probabilities of its constituent phonemes, which leads to smaller lexical items. Finally, the probability of generating an existing word is proportional to the number of times it has already occurred in the current segmentation, which leads to a power-law distribution over words similar to the distribution observed in natural languages \cite{zipf1932selected}  %
The bigram model is hierarchical (namely, a hierarchical Dirichlet process \cite{teh2006hierarchical}), and it achieves 72.3 token F-score, and 85.2 boundary F-score. The improvements are mainly due to increased recall, which is a result of breaking down collocations to their constituent word types. While this leads to much better segmentation, the bigram model does introduce another kind of error: over-segmentation of frequent word suffixes. %

Optimal word segmentation is likely to be achieved using interactive models that incorporate multiple levels of processing: words, collocations and sub-word structures like syllables or morphemes. One such interactive approach is the adaptor grammar described in \citet{johnson2008using}\footnote{An adaptor grammar is a probabilistic context-free grammar (PCFG) where some of the non-terminals and their probabilities are learned from data.}. The model learns hierarchical structures simultaneously by incorporating syllabic structure and collocations in addition to words in the grammar. Compared to word-only models, The highest improvement in word token F-score is achieved by specifying collocations in the grammar (0.76). A slight improvement is achieved by also incorporating syllabic structure (0.78), which is the highest word token F-score among all models discussed here. Modeling morphology in the form of stems and suffixes, however, did not improve performance.

\begin{table*}[t]
    \centering
    \begin{tabular}{|l|l|c|c|}
       \hline
       Model    &   Description & TF      & BF    \\
       \hline
        \citet{brent1999efficient} & probabilistic model with Unigram word distribution & 68 & - \\ 
       \citet{goldwater2009bayesian}  &  Non-parametric Baysian model with Unigram assumption & 54    & 74  \\ 
       \citet{goldwater2009bayesian}   & Non-parametric Baysian model with Bigram assumption  & 72    & 85  \\
        \citet{johnson2008using} & Adaptor grammar that models words only & 55 & - \\
       \citet{johnson2008using} & Adaptor grammar with words and Collocations & 76 & - \\
       \citet{johnson2008using} & Adaptor grammar with words, collocations \& syllables & 78 & - \\
       \citet{fleck2008lexicalized} & Unigram word distribution, incorporates pause information & 71 & 83 \\
       \citet{elsner2017speech} & Utterance autoencoder model with limited memory & 72 & 83 \\
       \hline
        \multicolumn{4}{|l|}{\cellcolor{black!10}\textit{Phonetically Transcribed Data \cite{elsner2012bootstrapping} }} \\
         \hline
       \citet{goldwater2009bayesian}   & Bigram probabilistic model & 62   & 80 \\ 
       \citet{elsner2013joint}   & Bigram probabilistic model with FST noisy channel & 67   & 82 \\
       \hline
    \end{tabular}
    \caption{A summary of word segmentation performance using token F-score (TF) and boundary F-score (BF) on the CHILDES dataset. Top, phonemically transcribed data\cite{brent1999efficient};  bottom, phonetically transcribed variant \cite{elsner2012bootstrapping} where a word could be transcribed differently in different contexts.}
    \label{tab:my_label}
\end{table*}

More recently, deep neural networks have been proposed for the task of segmentation to model human memory and lexical perception.  An unsupervised LSTM autoencoder with limited memory is explored in \citet{elsner2017speech}, optimized with the objective of minimizing utterance reconstruction errors. The boundary detection is optimized by sampling to estimate the gradient of the reconstruction loss: %
boundaries that appear in samples with low reconstruction loss are assigned a higher likelihood. Cross-entropy is used to estimate the probability of the data given a boundary. 
Limiting the memory may encourage the model to rely on phonological predictability within words and syntactic or semantic predictability between words. The intuition is that actual words should be easier to compress and reconstruct in the autoencoder model compared to random sequences of phonemes. Experiments show that networks comprised of a smaller number of hidden units outperform those with a larger number of hidden units.

While word segmentation is made easier by the unrealistic phonemic transcription in this dataset, natural speech actually contains other signals that could be exploited to aid segmentation. \citet{fleck2008lexicalized} is an example of a model that uses pause information to identify likely starts and ends of words. It corresponds to a unigram language model: it assumes words are independent given a word boundary. The pauses that occur naturally in spoken corpora are used to estimate the probability of a boundary given the left and right context around it. These probabilities are estimated using simple ngram statistics with backoff \footnote{The model is bootstrapped using a generous estimation of these probabilities: if a pause occurs at least once before or after a context, the probability is set to a high value close to 1.}.
The model performs similar to \citet{goldwater2009bayesian} for English, but worse for Spanish. Results indicate that the success of the model depends on the size of the corpus and the presence of pauses. It also generalizes well to Arabic\footnote{77 BF and 57 TF compared with \citet{goldwater2009bayesian}: 64 BF and 33 TF.}, a morphologically complex language with longer words. 

Raw speech contains variations in pronunciation, which are normalized in the dataset above using a phonemic dictionary. In \citet{elsner2012bootstrapping}, they construct an approximate phonetic transcription by converting each word randomly to one of the possible surface forms in the Buckeye corpus \cite{pitt2005buckeye}.  
Results on this dataset for \citet{goldwater2009bayesian} is 80.3 BF, and 62.4 TF. With noisy data, the model tends to over-segment. 
The accuracy of the model is improved by modeling phonetic variability explicitly using a noisy-channel model implemented as a finite-state transducer \cite{elsner2013joint}. The FST is optimized using the EM algorithm and initialized using faithfulness features to encourage plausible changes. The results of the Bigram model with the FST transducer optimized jointly is 81.5 BF, and 66.9 TF.

In \cite{jansen2013summary}, generative word segmentation models similar to the works described above (\cite{johnson2008using}) are evaluated on both phonemic transcriptions and automatically generated transcriptions using supervised and unsupervised models. Results confirm the conclusions reached by earlier models, namely that modelling syllables, words, and collocations together indeed improves the segmentation performance, even in the presence of noise as a result of using unsupervised acoustic models.

\subsection{Phonetic Segmentation of Raw Speech}\label{sec:phone}

Phones are distinct sounds produced by modifying the shape of the vocal tract, which is indirectly encoded in the spectral patterns of the speech signal. 
The speech signal is continuous, and phones vary according to the context of preceding and following phones (co-articulation), so identifying phonetic boundaries and clustering the segments into sets that correspond to actual phone categories is not an easy feat. However, spectral changes in the speech signal could be used to detect phonetic boundaries with high accuracy. 

Assuming that speech frames are more similar and predictable within than across phone boundaries, statistical models could be used to identify points of low predictability. In \citet{michel2017blind}, a simple pseudo-Markov model is proposed to estimate frame transition probabilities. %
An alternative LSTM model is also proposed, where the objective is to predict the next frame given past input. A peak prediction algorithm is then used to identify local maxima in prediction errors as potential phone boundaries. 

In \citet{lee2012nonparametric}, a generative Bayesian model is proposed to jointly segment speech into sub-word units that correspond to phones, cluster the segments into hypothesized phoneme types, and learn an HMM for each cluster. This generative model assumes phones are generated independently, and each phone is modeled as a three-state HMM. Each state's emission probability is modeled by a GMM with 8 components. This formulation roughly corresponds to standard acoustic models in traditional ASR systems, but employs an iterative inference process using Gibbs sampling to find the model that best represents the observed data.

\begin{table*}[t]
    \centering
    \begin{tabular}{|l|l|c|c|c|c|}
     \cline{3-6}
    \multicolumn{2}{c|}{} & \multicolumn{2}{|c|}{TIMIT} & \multicolumn{2}{|c|}{Buckeye} \\
    \hline
       Model  & Description & F1 & R-val & F1 & R-val\\
       \hline
       \citet{lee2012nonparametric}$\dagger$  & Bayesian acoustic model & 76.3 & 76.3 & - & - \\  
       \hline
       \citet{michel2017blind}  & Peaks in frame prediction errors & 78.2 & 80.1 & 67.8 & 72.1 \\
       \citet{wang2017gate} & Maxima in gate activation signals & - & 83.2 & 71.0 & 74.8 \\
       \citet{kreuk2020self} & contrastive learning & 83.7 & 86.0 & 76.3 & 79.7 \\
        \citet{baevski2021unsupervised} & k-means & 53.9 & 56.1 & - & - \\
        \citet{baevski2021unsupervised} & k-means + Viterbi & 62.9 & 66.5 & - & - \\
       \hline \hline
       \citet{kreuk2020phoneme} & SOTA supervised & 92.2 & 92.8 & 87.2 & 88.8 \\
               \hline
      
    \end{tabular}
    \caption{Phonetic segmentation boundary F1-scores and R-values on TIMIT and Buckeye datasets. The bottom row is a state-of-the-art supervised phoneme segmentation model for comparison. Table is reproduced from \citet{kreuk2020self} and \citet{baevski2021unsupervised} . $\dagger$ Results from original paper and on TIMIT training rather than test set.} 
    \label{tab:my_label}
\end{table*}

Using gated networks trained as frame autoencoders, Gate Activation Signals (GAS) could also be used for phonetic segmentation. \citet{wang2017gate} demonstrates that the temporal structure in these signals, particularly the update gate in Gated Recurrent Networks (GRNN), correlate with phone boundaries. The local maxima in these signals are used as potential boundaries for segmentation. 

Contrastive learning is a self-supervised learning framework where the objective is to group adjacent regions of the input together and to push disjoint regions away from each other \cite{jaiswal2021survey}. In \citet{kreuk2020self}, self-supervised contrastive learning is used to learn discriminative encodings of speech frames such that adjacent frames have higher cosine similarity than randomly sampled distractor frames\footnote{The function is implemented as a convolutional neural network.}. The learned encoding function is then used to detect phone boundaries at points where adjacent frames exceed a threshold dissimilarity value. A validation set was used to set the peak detection threshold and other hyperparameters using 10\% of each corpus. 
This model achieves state-of-the-art segmentation R-value (see table 2). The segmentation results can be generalized to out-of-domain data and other languages, especially if the training data is augmented with additional unlabeled speech. 

The model described in \citet{shain2020acquiring} encodes and decodes the speech signal in a hierarchical manner, where each layer encodes the input at different time scales, using a hierarchical multi-scale LSTM (HM-LSTM) \cite{chung2017hierarchical}. This model is an extension of the working memory model described in section \ref{sec:seg1}, but it includes a hierarchy of segmentation in different layers of the encoder, and the objective incorporates both memory (i.e. reconstruction) and prediction components. In experiments, however, the model did not work beyond phonetic segmentation in the first layer \footnote{In ZRC'15 dataset, the first layer achieves phone boundary F-score of 49.3 for English, and 53.8 for Xitsonga, much lower than the scores in Table 2.}.
Results also indicate that both memory and prediction pressures lead to balanced precision and recall ratios, where memory pressure slows down the segmentation rate and prediction pressures increase it, resulting in a more balanced precision and recall trade-off. 

The state-of-the-art unsupervised speech recognition model recently described in \cite{baevski2021unsupervised} employs a rather simple phonetic segmentation approach based on frame-wise embedding and clustering. All frame embeddings are first clustered using k-means, and then phone boundaries are initialized at points where the cluster ID changes. This simple approach results in boundary F-score of 54 on TIMIT. The segmentation is improved using Viterbi decoding after classifying the segments using the proposed unsupervised model (described in more details in section \ref{sec:dist}).

\subsection{Syllabic Segmentation from Raw Speech} \label{sec:syl}

Phones are linguistically well-defined and consistently transcribed in many datasets, but the syllable is often considered a better candidate for a basic unit in human speech perception \cite{port2007words,rasanen2012computational}. 
One advantage of using raw speech is the ability to identify the rhythmic patterns of syllables which is absent in phonetically transcribed inputs. 
Prosodic cues, like stress patterns in English speech, can be correlated with word boundaries. It has been estimated that about 90\% of words in spoken English begin with strong syllables \cite{cutler1987predominance}, 
and experiments suggest that infants younger than 10 months are more likely to segment words at the onset of strong syllables \cite{jusczyk1999beginnings}. Although not all languages have consistent stress patterns \cite{hyman1977nature}, the onset of syllables could be used to constrain the locations of word boundaries in combination with other statistical methods.

While syllables are not clearly defined and may overlap in time \cite{villing2006performance, goslin1999syllable}, their rhythmic structure can be identified using the acoustic features of speech \cite{rasanen2018pre}. In \citet{rasanen2015unsupervised}, unsupervised segmentation of syllables is used as the first step in word segmentation. A syllable is defined here as a segment of speech characterized by rhythmic increase or decrease of the signal's amplitude within 2-10 Hz. The waveform envelope (how the amplitude changes over time) is the main feature used for syllable boundary detection in several earlier models\footnote{See \citet{villing2006performance} for a review and comparison of these methods.}  \cite{mermelstein1975automatic, villing2004automatic, wu1997integrating}.

In \citet{rasanen2015unsupervised}, a damped harmonic oscillator driven by the speech envelope is proposed as an alternative syllabic segmentation algorithm. This algorithm is inspired by models of human neuronal oscillations assumed to be responsible for speech perception \cite{giraud2012cortical}. Unlike previous models that directly use the peaks and troughs of the amplitude envelope to identify syllables, this model uses the speech envelope to feed the oscillator. The oscillator's minima are then marked as potential syllable boundaries. 

After identifying syllabic segments, each segment is compressed into a fixed-dimensional vector by averaging their MFCC vectors\footnote{In \citet{rasanen2015unsupervised}, they divide each segment into 5 uniform parts and average the MFCC vectors in those sub-segments. The average vectors are then concatenated to get a fixed-length representation of each syllable.}. After segmentation and compression, the standard k-means algorithm is used to cluster the syllables (alternative non-parametric models for syllable clustering are explored in \cite{seshadri2017comparison}). Finally, recurring syllable sequences (n-grams of different orders) are identified as words. Since the syllable embeddings are based on MFCC features, the model is speaker-dependent, and the processing is done separately for each speaker. Compared to other syllable segmentation models, using the oscillator-based algorithm results in better word segmentation performance in the Buckeye corpus as shown in table \ref{tab:my_label}.

\subsection{Word Segmentation from Raw Speech}\label{sec:word}

The models described above for phone and syllable segmentation could be used as the first steps to achieving word segmentation using techniques similar to those described in section \ref{sec:seg1}. One example of that is the syllabic segmentation model described in the previous section, which is used with n-gram modeling to identify recurring syllables as words (SylSeg in table \ref{tab:my_label}). More sophisticated segmentation models that incorporate collations and other features could potentially be used for the same purpose, but this territory has not been fully explored in the literature.

The challenge in word segmentation from raw speech is that words do not have an acoustic signature that could be used to estimate word boundaries. While syllables have identifiable rhythmic patterns and phones exhibit some internal coherence, words are rather arbitrary. Most word segmentation models incorporate some prior assumptions about words, such as minimum length, maximum number of syllables, or number of word types. Some models are based on the idea of reconstruction loss via autoencoders, such as the segmental audio word2vec \cite{wang2018segmental} and working memory model \cite{elsner2017speech}, and others are based on optimizing word clustering while segmenting the input \cite{kamper2017segmental,kamper2017embedded}. Except for \citet{elsner2017speech}, these models do not incorporate word bigram dependencies. 

Reconstruction-based models (i.e. autoencoders) rely on the assumption that sequences of phones or acoustic features that constitute words should be easier to reconstruct than other arbitrary sequences. 
The segmental audio word2vec model \cite{wang2018segmental} is an RNN sequence-to-sequence autoencoder trained jointly with a binary segmentation gate, which is optimized using reinforcement learning. The encoder and decoder are reset at segment boundaries, so each segment is reconstructed independently, and the rewards are calculated by penalizing reconstruction errors and the number of segments to avoid over-segmentation. The training objective of the autoencoder itself is equivalent to the audio word2vec model described in section \ref{sec:emb}, where each word is treated independently by resetting the encoder and decoder at segment boundaries. The segmentation gate and autoencoder are trained in iterations, fixing one to update the other. This model does not incorporate any constraints or assumptions about words other than the penalty on the number of segments, which could be tuned to achieve phonemic rather than word segmentation. On TIMIT, this model achieves a word boundary F-measure of 43, with higher recall than precision (52 and 37, respectively). 

The working memory model \cite{elsner2017speech} described in  section \ref{sec:seg1} can also be used for raw speech segmentation. This is an utterance auto-encoder, which sequentially embeds segments and utterances and then reconstructs the whole utterance segment by segment. Unlike the segmental audio word2vec where words are reconstructed independently, this model implicitly incorporates word-to-word dependencies by auto-encoding full utterances rather than individual words, and results in better segmentation performance \footnote{The two models were not compared directly on the same dataset, but the difference is large enough to support this conclusion (51 vs. 43 boundary F-score)}. For acoustic input, the Mean Squared Error (MSE) is used instead of cross-entropy as a loss function. In addition, a one-letter penalty is added to discourage very short segments, and the boundaries are initialized using automatic voice activity detection (VAD). Additional assumptions are incorporated by limiting the number of words per utterance and frames per word to 16 and 100, respectively. Experimental results indicate that memory limitations---in the form of dropout or smaller hidden layers---are useful for word boundary detection.

\begin{table*}[]
    \centering
    \begin{tabular}{|l|c|c|c|c|}
    \cline{2-5}
    \multicolumn{1}{c|}{} & \multicolumn{2}{|c|}{English} & \multicolumn{2}{|c|}{Xitsonga} \\
    \hline
         Model  &  Boundary F & Token F  & Boundary F & Token F\\
         \hline
         SylSeg \cite{rasanen2015unsupervised}  &  55.2 & 12.4 & 33.4 & 2.7 \\
         BES-GMM \cite{kamper2017segmental} &  62.2 & 17.9 & 43.1 & 4.0\\
         ES-KMeans \cite{kamper2017embedded} & 62.2 & 18.1 & 42.1 & 3.7 \\
         Working Memory \cite{elsner2017speech} & 51.1 & 9.3 & - & - \\
         \hline
    \end{tabular}
    \caption{ Speaker-dependent evaluation of word segmentation models using word boundary and token F-scores. Results are reproduced from \cite{kamper2017embedded} except for the working memory model, which is added from their original paper. The evaluation is performed on ZRC'15 dataset, Left: English, Right: Xitsonga. Results are calculated per speaker, then averaged.  }
    \label{tab:my_label}
\end{table*}

The problem of segmentation is intertwined with the problem of clustering: identifying which segments are realizations of the same underlying type. The repeated occurrence of patterns is a crucial ingredient in statistical modeling. In speech, however, each occurrence of a word type in an utterance is somehow unique due to the variable nature of speech. Models that combine segmentation and clustering \cite{kamper2017segmental,kamper2017embedded} can improve the chance of identifying these repeated patterns in a large corpus. 

The general idea behind these joint models is to optimize both segmentation and clustering in iterations: given an initial set of word boundaries (e.g. uniform or syllabic boundaries), the segments are embedded into fixed-dimensional vectors and clustered into K-word types. Given this clustering, the segmentation is updated and optimized for each utterance. The process repeats thus in iterations until convergence. These models could potentially incorporate prior assumptions about words, such as the number of word types in the lexicon (i.e. number of clusters), the maximum number of syllables per word, and minimum word length. 

The Bayesian Segmental GMM (BES-GMM) \cite{kamper2017segmental}, jointly segments and clusters the input into hypothesized word types using a Bayesian GMM, where each mixture component corresponds to a word type. The GMM model can be viewed as a whole-word acoustic model: it defines a probability distribution over words in the lexicon. The segmentation and clustering are carried out iteratively: given a random initial segmentation, the GMM is used to cluster the segment embeddings; given the current GMM, a dynamic programming Gibbs sampling algorithm is used to find high-probability segments based on the current acoustic model. And so on until convergence. 

The embedded segmental K-means (ES-KMeans) \cite{kamper2017embedded} is an approximation of the BES-GMM model. Instead of Baysian inference, the segments are clustered using the standard k-means algorithm. In speaker-dependent evaluation (table 4), ES-KMeans achieves similar word segmentation performance as BES-GMM while being faster and more scalable. 
Like BES-GMM, ES-KMeans alternates between segmentation and clustering to jointly optimize the cluster assignments and segmentation. The objective function optimizes the segmentation and cluster assignments jointly, where the clustering part is the same as the K-means objective: minimizing the sum of square distances between segment embeddings and their cluster means, weighed by the segment's duration (so the model prefers smaller segments). To bootstrap the process, word boundaries are initialized randomly. The segments are then embedded into fixed-dimensional vectors using down-sampling, and clustered with k-means. Given a clustering, the objective is reduced to utterance-wise minimization of the squared distances between each segment and the cluster mean, which is optimized using the Viterbi algorithm. 

To obtain the results in table 4, additional constraints were used for both BES-GMM and ES-KMeans to limit the possible word boundaries: the oscillator-based syllable segmentation algorithm (section \ref{sec:syl}) is used to eliminate unlikely word boundaries, where each word can have a maximum of six syllables. Additional improvements are achieved by also specifying a minimum duration of 200 ms. The fixed-length embeddings are obtained by down-sampling. These features are not robust to speaker variations, which is why the models are evaluated in a speaker-dependent settings. In \citet{kamper2017segmental}, they also experiment with speaker-independent features following the approach in \citet{kamper2015unsupervised}, but the improvements are mediocre. Using additional data with more speakers, the ES-KMeans model can be used as a speaker-independent model; it achieved 52.7 boundary F-score and 13.5 token F-score in speaker-independent evaluation. 

While joint clustering models surpassed other existing models in unsupervised word segmentation,  they still suffer from over-segmentation---probably in part due to their simple unigram assumption. Basically, these models do not consider word-to-word transition probability; they only find likely segments that overlap in their acoustic features. Qualitative assessment of the clusters show that they contain acoustically similar segments, even if they do not map to the same word label \cite{kamper2017embedded}.

\section{Cross-Modal Alignment}\label{sec:dist}

In the above sections, we described methods for automatically segmenting and clustering audio signals into phones, syllables, or words. What remains is discovering the actual identity of those segments (i.e. classification). Without direct supervision in the form of transcribed speech, classification could potentially be achieved using unsupervised alignment techniques similar to those applied in the text domain for unsupervised cross-lingual word mapping \cite{lample2018word, aldarmaki2018unsupervised, artetxe2018robust}. Typically, this is achieved using Generative Adversarial Networks (GANs) to map sequences from the source to the target space; for ASR, this corresponds to mapping speech to text segments using unrelated speech and text corpora---in what we refer to as distant supervision.

Note that most proposed models that claim to be completely unsupervised use some form of rudimentary supervision, either oracle segment boundaries, or a labeled validation set. Without any form of supervision, no model achieved remarkable accuracy until very recently, where an unsupervised ASR framework was proposed that managed to significantly improve performance without any form of supervision \cite{baevski2021unsupervised} . 
The majority of models operate at the sub-word level, where a phoneme classifier is trained via distant supervision using phonemized text. These models are discussed in section \ref{sec:phonemap}. A small subset of models operate at the word level, where other forms of grounding, such as raw text or images, are used. We briefly discuss these in section \ref{sec:ground}. 

\subsection{Phonetic Alignment}\label{sec:phonemap}

For phones, distant supervision for cross-modal alignment has been explored in \citet{liu2018completely}. Given a sequence of phone segments (which, theoretically, can be acquired in an unsupervised fashion), the segments are embedded into fixed-length vectors, and the standard K-means algorithm is used to cluster those embeddings. The cluster sequences are then used as input to the GAN in order to map each cluster to a specific phoneme. The segmentation and clustering are done first, then the GAN is trained on the cluster sequences and true phoneme sequences from an independent text corpus\footnote{A lexicon is used to transform the text corpus to phoneme sequences.}.

A similar idea is employed in \citet{chen2019completely} and \citet{yeh2018unsupervised}, where a GAN is used to learn a phoneme-based unsupervised ASR. However, in these models, the segmentation and mapping are learned jointly in an iterative manner: after an initial segmentation\footnote{The initial segmentation is obtained using the unsupervised phoneme segmentation approach described in \citet{wang2017gate}.}, (1) a frame-level phoneme classifier is trained by matching the distribution of an independent phoneme language model, and (2) the phoneme boundaries are updated using the learned classifier. These two steps are repeated iteratively until convergence. 

In \citet{chen2019completely}, the phoneme classifier is learned using a GAN, then the GAN-generated labels are used to train phoneme HMMs to update the boundaries by force alignment. In \citet{yeh2018unsupervised}, the phoneme classifier is trained using Empirical Output Distribution Matching (Emperical-ODM) \cite{liu2017unsupervised} to match the distribution of the independent language model, and the boundaries are updated using simple MAP estimation. HMMs are used as a final step to refine the model. Both models incorporate an intra-segment loss to ensure that frames within a segment have similar output distributions (to model the internal consistency of phonemes). A comparison of these models is shown in Table 5. 

\begin{table*}[]
    \centering
    \begin{tabular}{|l|c|c|}
    \hline
    Model & Matched PER & Non-Matched PER \\
    \hline
    \multicolumn{3}{|l|}{\cellcolor{black!10}\textit{Supervised}} \\
    \hline
    Phoneme classifier &  28.9&  -\\
    \hline
    RNN transducer &  17.7 & -\\
    \hline
    \multicolumn{3}{|l|}{\cellcolor{black!10}\textit{Unsupervised w. Oracle boundaries}} \\
    \hline
    cluster GAN \cite{liu2018completely} & 40.2 & 43.4 \\
    \hline
    Segmental empirical ODM  \cite{yeh2018unsupervised} & 32.5 & 40.1 \\
    \hline
    phoneme classifier GAN \cite{chen2019completely} & 28.5 & 34.3 \\
    \hline
    \multicolumn{3}{|l|}{\cellcolor{black!10}\textit{Unsupervised}} \\
    \hline
    Segmental empirical ODM + MAP \cite{yeh2018unsupervised} & 36.5 & 41.6 \\
    \hline
    phoneme classifier GAN + HMM \cite{chen2019completely} & 26.1 & 33.1 \\
    \hline
    Unsupervised wav2vec \cite{baevski2021unsupervised} & 16.6 & 24.4 \\
    \hline
    Unsupervised wav2vec + self-training \cite{baevski2021unsupervised} & 11.3 & 18.6 \\
    \hline
    \end{tabular}
    
    \caption{Cross-modal alignment results as phoneme error rate (PER). Results are reproduced from \cite{chen2019completely} and \cite{baevski2021unsupervised}.  Matched refers to the setting where text and speech are extracted from the same subset of TIMIT train, whereas in the non-matched setting different subsets are used for speech and text.  }
    \label{tab:my_label}
\end{table*}

Recently, an unsupervised model based on wav2vec 2.0 \cite{baevski2020wav2vec} has been proposed for fully unsupervised ASR \cite{baevski2021unsupervised}. The model also employs a GAN for phoneme mapping similar to the approaches above, using phonemized text for distant supervision. In addition to embedding the segments using the wav2vec 2.0 framework, PCA is used to retain the most salient features and mean-pooling for obtaining fixed-size embeddings for each segment. A silence label is added to the list of possible phonemes (and randomly inserted in the phonemized text for consistency), which results in significant performance gains. After GAN training, self training is used to refine the model using HMMs or fine-tuning of the original wav2vec model to improve the segmentation as well\footnote{Refer to the paper for additional implementation details, including a proposed automatic cross-validation metric for model selection.}. This model achieves state-of-the-art unsupervised ASR performance, significantly outperforming previously proposed models. The robustness of this model has been recently investigated in \citet{lin2021analyzing}, where they conduct experiments using different speech and text corpora. The lowest error rate is achieved when large amounts of both speech and text drawn from similar domains are used for training. Domain mismatch and spontaneous speech are the main factors that degrade the performance of unsupervised ASR, and could be mitigated to some extent by pre-training and increasing the amount of data used for training. 

\subsection{Semantic Alignment}\label{sec:ground}

\citet{chung2018unsupervised} explores the unsupervised cross-modal alignment of speech and text embeddings using semantic embeddings obtained by speech2vec \cite{chung2018speech2vec}, which are equivalent to the text-based word2vec, as described in section \ref{sec:emb}. The mapping is evaluated using oracle boundaries and automatic segmentation methods using BES-GMM, K-means, and Syllseg (see section \ref{sec:seg}). After embedding, the k-means algorithm is used to cluster the segments into potential word types. The mean of each cluster is used as the unique embedding for the word type represented by that cluster. After that, domain-adversarial training, similar to a popular approach used in cross-lingual mapping of word embeddings \cite{lample2018word}, is used to map the word embeddings from the speech to the text domain (the training is similar to GANs).  The mapping is evaluated on a task related to ASR, which is spoken word classification, but it results in very low accuracy. However, since the embeddings have semantic features, the spoken segments are often mapped to semantically related words, consistent with the behavior in cross-lingual word mapping. In spoken word synonyms retrieval, the model achieves 57\% precision@5 on English using true word boundaries and identities, compared to 67\% using a dictionary for alignment. Using BES-GMM and k-means (i.e. completely unsupervised setting), the performance drops to 37\% as a result of segmentation errors.  Similar performance is achieved on the spoken word translation task.


Another form of semantic grounding is using images to guide spoken term discovery \cite{harwath2015deep, harwath2016unsupervised, chrupala2017representations, harwath2019towards} and visually-grounded language modeling \cite{dunbar2021interspeech}. However, these models rely on aligned image-caption pairs, and they can only be used in unsupervised ASR if such alignments are available for text captions as well as audio captions. One potential application of visually-grounded acoustic models is in phonetic segmentation as recent analysis of activations show some diphone structure in multimodal neural models \citet{harwath2019towards}.

\section{Summary and Discussion}

We reviewed various research efforts in the direction of unsupervised speech recognition, including unsupervised sub-word modeling, spoken word embeddings, unsupervised term discovery, full-coverage segmentation, and cross-modal alignment. In this section, we summarize the main takeaways from this review and outline some of the challenges and possible directions for future research.  

The first three sub-tasks: sub-word and word-level feature representation, spoken term discovery, and segmentation, are often approached concurrently, particularly in more recent models, as they have overlapping objectives. Full lexical segmentation, for example, followed by acoustic word embedding and clustering, essentially amounts to full-coverage spoken term discovery. However, approaching the sub-tasks individually can also have advantages leading to better performance in the sub-tasks, and subsequently, better performance in the overall unsupervised ASR pipeline. For instance, as observed in the few cases where a segmentation model has been evaluated in both spoken term discovery and full-coverage segmentation (see,  for example, \citet{kamper2017embedded}), the full-coverage segmentation variants tend to have lower precision compared with models that attempt to directly discover recurring terms. Also, the state-of-the-art unsupervised ASR model described in \citet{baevski2021unsupervised} has lower phonetic segmentation accuracy compared to models that specialize in phonetic segmentation, such as \citet{kreuk2020self}. The state-of-the-art could potentially be improved by incorporating the best practices in each sub-task for initialization or fine-tuning. 

In sub-word and word modeling, the purpose is to finds suitable representations that downplay irrelevant features such as speaker-specific characteristics, while emphasizing features that distinguish between different acoustic units; this can be carried out before segmentation at the level of frames, which can essentially lead to acoustic unit discovery, or after segmentation at the level of sub-word or word segments. As shown in more recent iterations of the Zero Resource Speech challenge, using Contrastive Predictive Coding (CPC) leads to more robust frame-level features that are better in distinguishing phonetic categories and are somewhat speaker-invariant \cite{dunbar2020zero}. In fact, contrastive learning has been shown repeatedly to be a superior sub-word modeling method. In phonetic segmentation, the state-of-the-art model \cite{kreuk2020self} uses contrastive learning to learn frame-wise features, and phonetic boundaries are inserted at points where adjacent frames exceed a dissimilarity threshold. CPC has also been shown to lead to better acoustic word embeddings compared with MFCC raw features \cite{van2021comparison}. 

Efforts in unsupervised speech segmentation include phonetic, syllabic, and lexical segmentation. Phonetic and syllabic segmentation are more manageable than word segmentation, as they can be obtained directly by analyzing the characteristics of the speech signal. The best performing models in unsupervised phonetic segmentation, for example, achieve a boundary F-score above 80. On the other hand, the best word segmentation from raw speech achieve a boundary F-score around 60, and less than 20 token F-score. Earlier efforts in word segmentation from phonemically-transcribed speech indicate that better results could be obtained by modeling bigrams in addition to individual words, or incorporating additional features such as pauses to identify boundaries more robustly. Yet, most recent works on word segmentation from raw speech do not model word dependencies. Explicit modeling of collocations, in addition to incorporating pauses and utterance boundaries, could potentially improve performance in these models. In addition, while syllabic segmentation has been incorporated in some lexical segmentation models to constrain word boundaries around the onset of syllables, the higher success of phoneme segmentation models could be used to construct lexical models that incorporate the phoneme as a building block. 

In addition to segmentation, the choice of embedding and clustering methodology is important for unsupervised ASR. Embeddings that emphasize semantic features can be useful in unsupervised speech translation or query-by-example search, but they are insufficient to accurately label spoken words for automatic speech transcription. On the other hand, embeddings that favor phonetic features are more suitable for this task, but they are harder to align automatically with text-based embeddings. Clustering based on phonetic embeddings could lead to spoken term discovery, but due to variability in spoken terms, the discovered clusters are often speaker-dependent. Some efforts in this vein, such as using large datasets with multiple speakers or using features that attempt to isolate phonetic from speaker-specific features, led to minor improvements in speaker-independent evaluation, but there is still large room for improvement to make unsupervised methods robust to speaker variations.  

Distant supervision using Generative Adversarial Networks has been explored recently for mapping speech segments to corresponding text segments, using both phones and words as segmental units. Compared to word-based models, phone mapping has shown better promise, with error rates below 45\%. The state-of-the-art model in this category achieves remarkable success (around 11\% word error rate in the given benchmark) by incorporating a GAN for phone mapping in addition to refining the segmentation and embeddings using self-training. Word-level cross-modal mapping, on the other hand, has only managed to retrieve semantically related words, which could be useful in translation tasks, but not in ASR where exact matches are required. A potential future development could involve the combination of phone and word mapping, where phonetic mappings provide bottom-up labels, while semantic lexical mapping provide a top-down signal to constrain and improve the lower-level alignments. Recent unsupervised ASR models that achieve encouraging results \cite{baevski2021unsupervised} still operate at the phonetic level, and so they require phonetic or phonemic text transcriptions for cross-modal mapping. Operating at the word level, on the other hand, would make it possible to align speech with raw text; however, such models rely on the lexical segmentation accuracy and may require acoustic word embeddings that encode semantic features such as speech2vec (see, for example, \citet{chung2018unsupervised}). 

A sizable gap still exists between supervised, semi-supervised, and unsupervised models, but recent efforts show that closing that gap is not only possible, but could just be a matter of finding the right combination of existing strategies to achieve optimal performance.

\section*{Acknowledgement:} This work was supported by grant no. 31T139 at United Arab Emirates University. 

\bibliography{sample}
\bibstyle{acl_natbib}

\end{document}